%% file: main.tex
\documentclass[letterpaper, 10 pt, conference]{ieeeconf}  

\IEEEoverridecommandlockouts     

\usepackage{subcaption}
\usepackage{graphicx} 

\usepackage{epsfig} 
\usepackage{mathptmx} 
\usepackage{times} 
\usepackage{amsmath} 
\usepackage{amssymb}  
\usepackage{xcolor}
\usepackage{tabularx}
\usepackage[ruled,vlined,linesnumbered]{algorithm2e}  
\usepackage{lipsum}  
\usepackage{multirow}
\usepackage{booktabs}
\usepackage{caption}
\usepackage{hyperref}

\newtheorem{definition}{Definition}

\usepackage{tabularray}

\title{\LARGE \bf
A Game-Theoretic Framework for Joint Forecasting and Planning
}

\author{Kushal Kedia$^{1}$, Prithwish Dan$^{1}$, Sanjiban Choudhury$^{1}$
\thanks{$^{1}$The authors are affiliated with Cornell University, Ithaca, New York, USA
        {\tt\small \{kk837, pd337, sc2582\}@cornell.edu}}%
}

\begin{document}

\maketitle
\thispagestyle{empty}
\pagestyle{empty}

\begin{abstract}
Planning safe robot motions in the presence of humans requires reliable forecasts of future human motion. However, simply predicting the most likely motion from prior interactions does not guarantee safety. Such forecasts fail to model the long tail of possible events, which are rarely observed in limited datasets. On the other hand, planning for worst-case motions leads to overtly conservative behavior and a ``frozen robot''. Instead, we aim to learn forecasts that \emph{predict counterfactuals that humans guard against}. We propose a novel game-theoretic framework for joint planning and forecasting with the payoff being the performance of the planner against the demonstrator, and present practical algorithms to train models in an end-to-end fashion. We demonstrate that our proposed algorithm results in safer plans in a crowd navigation simulator and real-world datasets of pedestrian motion.  We release our code at \url{https://github.com/portal-cornell/Game-Theoretic-Forecasting-Planning}.
\end{abstract}

\input{inputs/introduction.tex}
\input{inputs/related_work.tex}

\input{inputs/problem_formulation.tex}
\input{inputs/approach.tex}
\input{inputs/experiments.tex}
\input{inputs/discussion.tex}

\section{Acknowledgements}
This work was partially funded by NSF RI (\#2312956).
\bibliographystyle{IEEEtran}
\bibliography{references}
\end{document}

%% file: inputs/introduction.tex
\section{Introduction}
One of the greatest challenges in robotics and AI is reasoning about interaction with other agents in the world. The ability to forecast how other agents behave in response to a robot's decisions is key to enabling safe, interpretable, and responsive behavior. Consider a self-driving car negotiating a left turn at a busy intersection, or a personal robot collaboratively cooking with a human in a kitchen. The robot has to both yield to the human at times, and show intent to go ahead at other times. To do so, it must rely on forecasts that are conservative enough to predict rare but risky events, but not so conservative that the robot stays frozen in place. 


Current forecasting approaches are primarily based on Maximum Likelihood Estimate (MLE). For instance, in self-driving, state-of-the-art forecasters~\cite{ngiam2021scene, li2020end} are typically trained on the L2 loss between the observed future motions of actors and the predicted motion on data collected off-policy. 
A planner then uses the forecast to compute a safe, collision-free path.
However, a forecaster trained purely on observed data \textbf{fails to predict rare but risky events}. The distribution of motions exhibits a long tail, necessitating the modeling of this tail with exorbitant amounts of data to accurately represent the diverse rare events.

\begin{figure}[t!]
    \centering
    \includegraphics[width=\columnwidth]{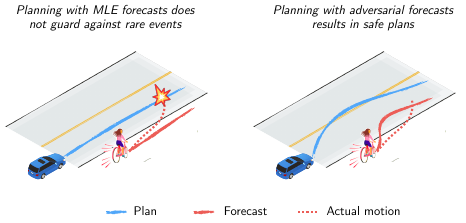}
    \caption{MLE forecasts fail to predict rare, hazardous events, like a bicyclist suddenly veering into a car's lane. We propose to learn adversarial forecasts that enable a planner to guard against such events.}
    \label{fig:motivation}
    \vspace{-4mm}
\end{figure}

Consider the example in Fig.~\ref{fig:motivation} of a self-driving car driving alongside a cyclist. We observe humans leave their lane to give space to a cyclist while actively occupying the lane of an oncoming car. However, an MLE forecaster will likely predict the cyclist staying in their lane. This results in plans that fail to guard against possible rare events such as the cyclist accidentally coming into the lane. An alternate approach is to reason about the worst-case outcome given the reachability of the cyclist~\cite{bajcsy2020robust, leung2021towards}. But this can lead to overtly conservative behavior where the robot stays perpetually behind the cyclist. 



\textbf{We propose a new objective for training forecasters.} We argue that MLE loss on observed motions from a finite dataset is fundamentally insufficient due to lack of coverage in the dataset. Instead, we view the problem through the lens of imitation learning. Our key insight is that \emph{humans don't just plan for things that are likely to happen, but plan contingencies for counterfactuals that could possibly happen}. We aim to learn forecasts that enable a planner to guard as well as the demonstrator against any possible rare event. We propose a game-theoretic framework for jointly training planners and forecasts, and use no-regret learning to solve for the approximate equilibrium of the game. Our key contributions are summarized as follows:


\begin{enumerate}
    \item A novel, game-theoretic framework for joint forecasting and planning that guarantees performance with respect to demonstrations. 
    \item Practical algorithms and architectures for joint forecasting and planning for multi-agent navigation.
    \item Empirical evaluation on a crowd navigation simulator and real-world pedestrian datasets.
\end{enumerate}

\begin{figure*}[h!]
    \centering
    \includegraphics[width=\textwidth]{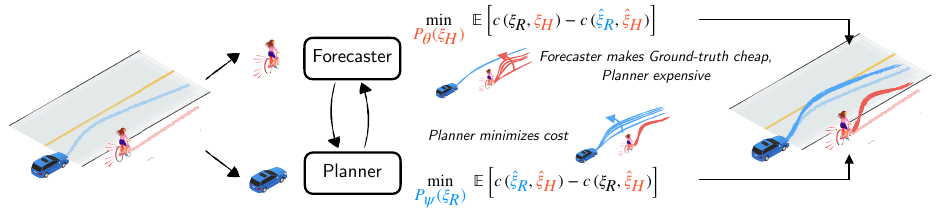}
    \caption{Overview of our game-theoretic framework for joint forecasting and planning. The forecaster maximizes the performance difference between the generated plans and the observed plans. This results in counterfactual forecasts for the cyclist veering into the vehicle's lane that encourages the planner to guard by nudging away from the cyclist.}
    \label{fig:approach}
    \vspace{-4mm}
\end{figure*} 

%% file: inputs/related_work.tex
\section{Related Work}
We focus on the problem of planning for robot decisions in the presence of humans in the workspace. The intended motions of the humans are unknown to the robot as it plans a sequence of actions that maximize the reward function for a given task. We look at various clusters of related work.


\textbf{Multi-agent games} 
Multi-agent games are formulated as both Stackelberg or Nash Equilibrium finding problems. \cite{Sadigh2016PlanningFA} models human action as a function of the robot's action and the system's current state. \cite{Sadigh2016InformationGA} extends this model to distinguish between different types of human actors (attentive and distracted).  \cite{FridovichKeil2020EfficientIL} solves for the Global Nash Equilibrium by using the Hamilton-Jacobi-Bellman equation of optimal control. Trajectory optimization utilized by \cite{Wang2021GameTheoreticPF} includes a sensitivity term that allows the vehicle to reason how much the other vehicles will yield to avoid collisions. To capture the problem's constraints effectively,  \cite{Cleach2022ALGAMESAF} enforces collision-avoidance through augmented Lagrangian constraints instead of imposing a large penalty on collision while constructing the objective functions for the problems. However, unlike our work, in order to solve these games, full information about the cost for each agent is required. 

\textbf{Autonomous Driving} In the autonomous driving domain, this problem is broken down into game-theoretic interactions to represent lane-merging, intersection crossing, pedestrian management, etc. Works in this domain have modeled the game by a Stackelberg equilibrium \cite{Liniger2020ANG, Sadigh2016InformationGA, Sadigh2016PlanningFA} where the behavior of a leader is fixed and best-response strategy is learned for the follower, or by a Nash equilibrium \cite{Cleach2022ALGAMESAF, FridovichKeil2020EfficientIL,Peters2021InferringOI,Wang2021GameTheoreticPF}, where agents follow strategies such that their objectives cannot be improved upon unilateral deviation. In this paper, we focus on the unstructured domain of pedestrian navigation, where there is large variability in human behavior and the space of possible motions is larger.  

\textbf{Forecasting Human Motion} Real-world human movements constitute a broad multimodal distribution containing inherent uncertainty and noise. Prior works have focused on effective model architectures and appropriate representations of interactions between agents. Human-robot and human-human interactions can be effectively modeled with a self-attention mechanism \cite{Chen2018CrowdRobotIC} for robot planning. Multimodal deep generative models \cite{Ivanovic2020MultimodalDG} for trajectory forecasting can effectively represent diverse human behavior. Trajectron++ \cite{Salzmann2020TrajectronDT} produces dynamically-feasible predictions by incorporating dynamics constraints into learned multi-agent trajectory forecasting. Representation learning \cite{Liu2020SocialNC} of safer motion representations can be facilitated by contrastive estimation from simulated negative behavior. The problem of transfer-learning forecasting models from different datasets has been tackled by explicitly modeling styles and noise confounders \cite{Liu2021TowardsRA}. However, the focus of forecasting literature has been largely restricted to task-agnostic accuracy-based metrics such as average displacement error (ADE) and final displacement error (FDE). For instance, in self-driving, it is important to use task-specific metrics prioritizing the safety of a planner that consumes the forecasts \cite{Philion2020LearningTE}. This has motivated the community to rethink appropriate metrics for planner-centric evaluation of forecasting \cite{Ivanovic2021RethinkingTF}. In this work, we are ultimately interested in generating forecasts that optimize the final performance of our downstream planner.

%% file: inputs/problem_formulation.tex
\section{Problem Formulation}
\label{sec:prob_form}
We model the forecasting problem as a multi-agent Contextual Markov Decision Process (CMDP), where one of the agents is the robot, and the rest are humans. Let $\phi$ denote the context -- a history of past states-actions for all the agents and the current scene. We assume contexts are drawn from a distribution $P(\phi)$. Let $\xi_R = (s^R_1, a^R_1, s^R_2, a^R_2, \dots, s^R_T, a^R_T)$ be the $T$-horizon trajectory of the robot. Let $\xi^H = (s^H_1, a^H_1, s^H_2, a^H_2, \dots, s^H_T, a^H_T)$ be the trajectory of all the other human agents. Let $c(\xi^R, \xi^H)$ denote the cost of a robot-human trajectory pair. This captures terms like safety, burden and deviation from the nominal path. 

We assume access to demonstrations of human and robot trajectories drawn from a joint probability $P(\xi_R, \xi_H | \phi)$. We aim to learn a conditional distribution over robot trajectories $P_\psi(\xi_R | \xi_H, \phi)$, where $\psi$ are the learnt parameters, that minimize the average performance difference with respect to the demonstrator.

\begin{equation}
    \label{eq:perf_diff}
    \begin{aligned}
     \mathbb{E}_{\phi} \left[ \underset{\substack{\xi_H \sim P(.|\phi) \\ \hat{\xi}_R \sim P_{\psi}(.|\xi_H, \phi) } }{\mathbb{E}} c(\hat{\xi}_R, \xi_H) - \underset{\xi_H, \xi_R \sim P(.|\phi) }{\mathbb{E}} c(\xi_R, \xi_H) \right]
    \end{aligned}
\end{equation}


\subsection{Pitfalls of Planning with MLE forecast}
\label{sec:prob_form:pitfalls}
A template for solving Eq.\ref{eq:perf_diff} is to first train a \emph{forecaster} to approximate the human trajectory distribution $P_\theta(\xi_H | \phi) \approx P(\xi_H | \phi)$, where $\theta$ are learnt parameters. A common way is to train a Maximum Likelihood Estimator (MLE).

\begin{definition}[\textsc{MLE-Forecaster}]
Given a dataset $\mathcal{D} = \{(\phi, \xi_H) \}$ of context and human trajectories, the goal of the MLE forecaster is to maximize the likelihood of observed trajectories:
\begin{equation}
    \label{eq:mle_forecasting}
    \max_\theta  \;\mathbb{E}_{\phi, \xi_H} \log P_\theta(\xi_H | \phi)
\end{equation}
\end{definition}

A nominal approach to planning is to minimize cost with respect to this learnt forecast. 
\begin{definition}[\textsc{Nom-Planner}]
Given access to a  \textsc{MLE-Forecaster} $P_\theta(\xi_H | \phi)$, a nominal planner minimizes expected costs w.r.t. the forecasts
\begin{equation}
    \label{eq:nominal_planning}
    \min_\psi \; \mathbb{E}_{\phi} \underset{\substack{\hat{\xi}_H \sim P_\theta(.|\phi) \\ \hat{\xi}_R \sim P_{\psi}(.|\hat{\xi}_H, \phi) }}{\mathbb{E}} c(\hat{\xi}_R, \hat{\xi}_H)
\end{equation}
\end{definition}

While \textsc{MLE-Forecaster}+\textsc{Nom-Planner} is industry standard, the framework has two fundamental problems:
  \begin{enumerate}[wide, labelwidth=!, labelindent=0pt]
    \item \emph{Failure to predict rare but risky events:} The MLE loss is dominated by events that occur frequently in the data. It fails to predict events $\xi_H$ that occur rarely in the data. However, these events can be quite risky, i.e., even if $P(\xi_H | \phi)$ is small, the cost $c(\xi_R, \xi_H)$ of the planner can be very large.  
    \item \emph{Small forecasting errors lead to large planning errors}: The MLE loss optimizes the KL-Divergence $KL( P(\xi_H | \phi) || P_\theta(\xi_H | \phi) )$, which is mismatched from the performance difference in (\ref{eq:perf_diff}). Formally, a bounded KL divergence, implies a Total Variation (TV) distance bound of $\epsilon$ (Psinker's inequality). However, a small error in forecasting could result in an approximation error of $C_{max} \epsilon$ in the downstream planner's performance, where $C_{max}$ is the maximum cost of a trajectory.  
  \end{enumerate}





%% file: inputs/approach.tex
\section{Approach}
We present a novel game-theoretic framework for joint forecasting and planning. We also present a concrete application of the framework in a multi-agent navigation setting. 

\begin{figure*}[t!]
    \centering
    \includegraphics[width=\textwidth, height=6cm]{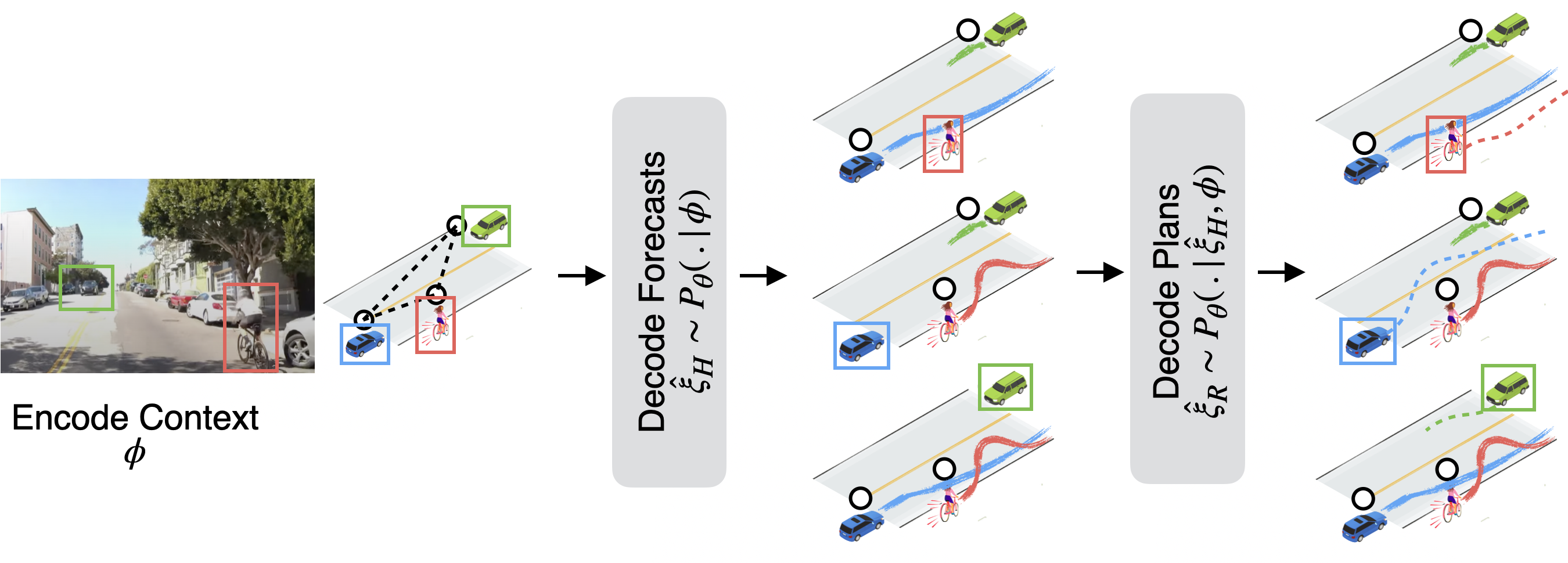}
    \caption{\textbf{Model architecture for joint forecasting and planning.} Agents in the scene are represented by nodes in the graph. Each agent has two outputs: a forecast and a plan. We apply first self-attention over the encoded contexts for each node which are decoded into adversarial forecasts, which are then used by the planner to generate safe plans for each agent.}
    \label{fig:model_architecture}
    \vspace{-6mm}
\end{figure*}

\input{inputs/approach/game_theoretic_framework}

\input{inputs/approach/training_details}

%% file: inputs/approach/game_theoretic_framework.tex
\subsection{Game-Theoretic Framework for Forecasting / Planning}
In Section~\ref{sec:prob_form:pitfalls}, we discussed two fundamental problems with the MLE approach: (1) failure to predict rare-events (2) loss mismatched with performance difference (Eq.\ref{eq:perf_diff}). We now propose an approach that addresses both problems. Our key insight is that \emph{humans don't just plan for things that are likely to happen, but plan contingencies for counterfactuals that could possibly happen}. For example, in Fig.~\ref{fig:motivation}, the human plans a path $\xi_R$ that guards for the counterfactual that the bicyclist may accidentally veer onto their lane. We aim to learn forecasts that \emph{don't just predict likely motions, but predict counterfactuals that humans guard against}.


We view the problem from the lens of inverse optimal control (IOC)~\cite{Ziebart2008MaximumEI}. IOC aims to learn a cost function that explains demonstrated behavior. Here, we aim to learn a forecast (that in turn defines the cost function) that explains demonstrated behavior. IOC in this setting can be best understood as a two-player zero-sum game~\cite{swamy2021moments} between a forecaster and a planner. The forecaster generates forecasts $\hat{\xi}^H$ that \emph{increase} the performance difference between the planner and the demonstrator. The planner generates plans $\hat{\xi}_R$ that \emph{decrease} the performance difference. Finally, to ensure that the forecasts are not completely unrealistic, we constrain them to be within an $\delta$-ball of the observed distribution. 
\vspace{-1mm}
\begin{equation}
    \begin{aligned}
    \max_{\theta} \min_{\psi} \; \mathbb{E}_{\phi} \; \left[
    \underset{\substack{\hat{\xi}_H \sim P_\theta(.|\phi) \\ \hat{\xi}_R \sim P_{\psi}(.|\hat{\xi}_H, \phi)} }{\mathbb{E}} c(\hat{\xi}_R, \hat{\xi}_H) - \underset{\substack{  \hat{\xi}_H \sim P_\theta(.|\phi) \\ \xi_R \sim P(.|\phi)}}{\mathbb{E}} c(\xi_R, \hat{\xi}_H) \right] \\
    \text{s.t.} \; \mathbb{E}_{\phi} [KL ( P_\theta(\hat{\xi}_H|\phi) \; || \; P(\xi_H|\phi) )] \leq \delta 
    \end{aligned}
    \label{eq:game_forecasting_planning}
\end{equation}

We aim to compute a (near-optimal) $\epsilon-$ equilibria of the game above, which would result in a planner $P_\psi$ that bounds the original objective (\ref{eq:perf_diff}) by $\epsilon$ as well. Following the arguments in~\cite{swamy2021moments}, since the game is bilinear in both $P_\theta$ and $P_\psi$, playing a no-regret strategy for both forecaster and the planner guarantees finding the $\epsilon-$ equilibria. 

\begin{algorithm}[t]
  \caption{ \textsc{Adv-Forecaster} / \textsc{Safe-Planner} }\label{alg:joint_forecaster_planner}
  \SetKwInOut{Input}{Input}
  \SetKwInOut{Output}{Output}

  \Input{ Dataset $\mathcal{D} = \{(\phi, \xi_R, \xi_H)\}$}
  \Output{ \textsc{Adv-Forecaster} $P_\theta(.|\phi)$, \textsc{Safe-Planner}  $P_\psi(.|\phi)$}
  Initialize $\theta_1$ with \textsc{MLE-Forecaster} \\
  Initialize $\psi_1$ with \textsc{Nom-Planner} \\
  \For{$i = 1 \dots N$}
  {
    Invoke current forecaster $P_{\theta_i}$ and planner $P_{\psi_i}$ on $\mathcal{D}$ to create a dataset $\{(\phi, \xi_H, \xi_R, \hat{\xi}^i_H, \hat{\xi}^i_R) \}$ \\
    Update planner $\psi_{i+1} \gets \psi_i - \nabla_\psi \ell^i(P_\psi)$ \\
    Update forecaster $\theta_{i+1} \gets \theta_i - \nabla_\theta \ell^i(P_\theta)$\\
  }
  \Return{$P_{\theta_N}(.|\phi)$, $P_{\psi_N}(.|\phi)$}
\end{algorithm}
We define the forecaster trained in this adversarial fashion as \textsc{Adv-Forecaster}, and the planner trained to be robust against such an adversarial forecaster as \textsc{Safe-Planner}. We describe the overall approach in Algorithm~\ref{alg:joint_forecaster_planner}. We setup an online learning game that lasts $N$ rounds. In every round, both the forecaster and the planner receive a loss function and play a no-regret update (we use online gradient descent). We define the loss functions for both players below: 

\begin{definition}[\textsc{Adv-Forecaster}] 
At round $i$, the forecaster receives a dataset $\{(\phi, \xi_H, \xi_R, \hat{\xi}^i_R) \}$ of context, human demonstration, robot demonstration, and the planned trajectory, respectively. We define the loss for this round $\ell^i(P_\theta)$ as :
\begin{equation}
    \begin{aligned}
    \ell^i(P_\theta) = \underset{\phi, \xi_H, \xi_R, \hat{\xi}^i_R}{\mathbb{E}} \left[ \underset{\hat{\xi}_H \sim P_\theta(.|\phi)}{\mathbb{E}} \left[c(\xi_R, \hat{\xi}_H) - c(\hat{\xi}^i_R, \hat{\xi}_H)\right] \right. \\
    \left. \vphantom{\underset{\hat{\xi}_H \sim P_\theta(.|\phi)}{\mathbb{E}}} -  \lambda \log P_\theta(\xi_H|\phi) \right] 
    \end{aligned}
    \label{eq:adv_forecasting}
\end{equation}
where the first term is the difference between the costs of the demonstrated and the planned robot trajectory, and the second term is the log-likelihood of observed human trajectories multiplied by a Lagrange multiplier. 
\end{definition}

\begin{definition}[\textsc{Safe-Planner}]
At round $i$, the planner receives a dataset $\{(\phi, \xi_R, \hat{\xi}^i_H) \}$ of context, robot demonstration and the adversarial forecast respectively. We define the loss for this round $\ell^i(P_\psi)$ as :
\begin{equation}
    \begin{aligned}
    \ell^i(P_\psi) = \underset{\phi, \xi_R, \hat{\xi}^i_H}{\mathbb{E}} \left[ \underset{\hat{\xi}_R \sim P_\psi(.|\hat{\xi}^i_H, \phi)}{\mathbb{E}} \left[ c(\hat{\xi}_R, \hat{\xi}^i_H) - c(\xi_R, \hat{\xi}^i_H)\right] \right]
    \end{aligned}
    \label{eq:safe_planning}
\end{equation}
where the inner most term is the difference between costs of the planned and the demonstrated robot trajectory against the current forecast.
\end{definition}

%% file: inputs/approach/training_details.tex
\subsection{Application for Multi-Agent Navigation}


We define a multi-agent navigation scene to contain $N$ agents interacting with each other. We can consider each agent, in turn, to act as the ``robot'' interacting with the other ``humans'' in the scene. For every agent, we have to plan a $T$-horizon trajectory, considering the future motions of the other agents in the scene.  We need to encode goal-reaching and collision avoidance to define a cost function for the navigation problem. However, while forecasting motions for agents, we do not always have access to their intended goal locations. From a dataset of prior interactions between agents, given the context of the scene, we can infer the future motions using maximum-likelihood estimation. Additionally, we enforce collision avoidance using the following obstacle cost function \cite{Zucker2013CHOMPCH} between plans and forecasts:
\begin{equation}
\begin{aligned}
    & c({\xi}_R, {\xi}_H) = \sum_{i}^{T} COL(s_{i}^{R}, s_{i}^{H}) \\
    & COL(s_{i}^{R}, s_{i}^{H}) = 
        \begin{cases}
          -dist(s_{i}^{R}, s_{i}^{H}) + \frac{1}{2} \epsilon, & dist(s_{i}^{R}, s_{i}^{H}) < 0 \\
          \frac{1}{2\epsilon}(dist(s_{i}^{R}, s_{i}^{H})-\epsilon)^2 & 0 < dist(s_{i}^{R}, s_{i}^{H}) < \epsilon \\
          0, & \text{otherwise}
        \end{cases}  
    \label{eq:chomp_cost}
\end{aligned}
 \end{equation}

We sum the cost over all robot and human states in the predicted $T$-horizon planner and forecaster trajectories. When a robot interacts with multiple humans, the human trajectory with the largest cost is considered.


In our model architecture (Fig. \ref{fig:model_architecture}), each agent in the scene is represented by a node in a graph. Neighboring agents are connected by edges. Each node in the graph takes in its individual context, including state history and other relevant information, such as a local map representation of the scene. To encode interactions with other agents, self-attention is applied across neighboring nodes. Each node has two output heads: a forecaster and a planner. The predicted forecasts can be fed as input to the planning module. While the inputs to the forecaster are restricted to the shared context of the scene, the planner can additionally take in information private to an agent, such as its goal location. 


We first train the forecasting and planning model to only maximize the likelihood of the future motion for the agents in the scene (Eq. \ref{eq:mle_forecasting} and Eq. \ref{eq:nominal_planning}). This is equivalent to simply maximizing the likelihood of the future motions in the dataset, and we call it a \textbf{\textsc{MLE-Forecaster}} and \textbf{\textsc{Nom-Planner}}. Apart from matching the ground truth, we wish to encode collision avoidance for measuring our plan's performance with respect to the forecasts. To incorporate this cost function, we solve the min-max game defined by Eq. \ref{eq:game_forecasting_planning}. For every minibatch of data, we update the forecasting model (\textbf{\textsc{Adv-Forecaster}}) to adversarially maximize the difference of the cost functions between the planner and the ground truth. In response, the planner (\textbf{\textsc{Safe-Planner}}) is updated to minimize the costs with the \textsc{Adv-Forecaster}. We ensure that the predictions do not deviate too much from the ground truth data by continuing to optimize the likelihood of the observed future motion.

%% file: inputs/experiments.tex
\section{Evaluation}
\input{inputs/figures/fig_costs}
\subsection{Setup}


We evaluate our algorithm on a crowd navigation simulator and real-world pedestrian datasets.

\input{inputs/experiments/crowdnav}

\input{inputs/experiments/eth_ucy}
\input{inputs/experiments/results}

%% file: inputs/figures/fig_costs.tex

\begin{figure*}[t!]
    \centering
    \begin{subfigure}[b]{.49\textwidth}
    \centering
    \includegraphics[width=.49\textwidth]{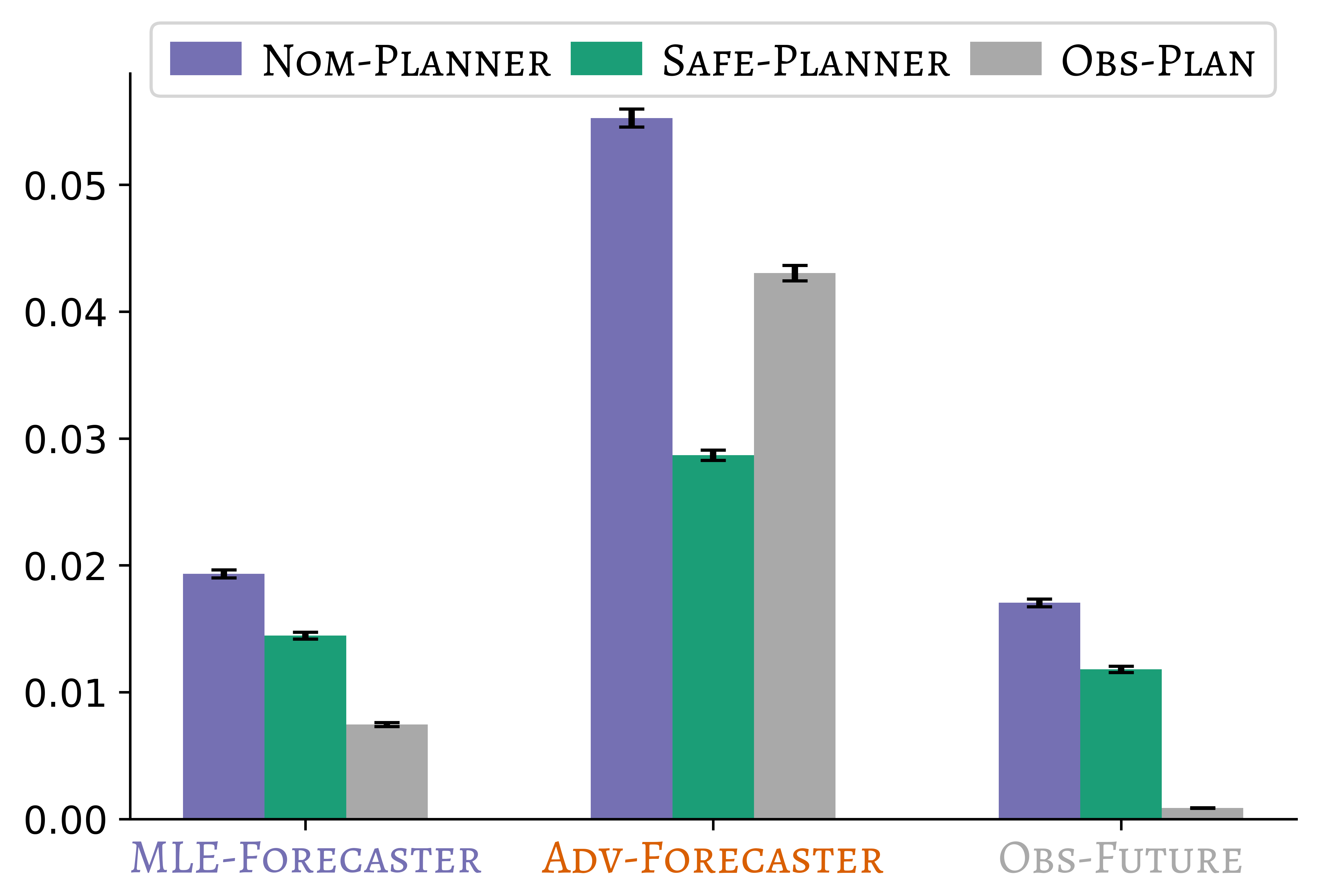}
    \includegraphics[width=.49\textwidth]{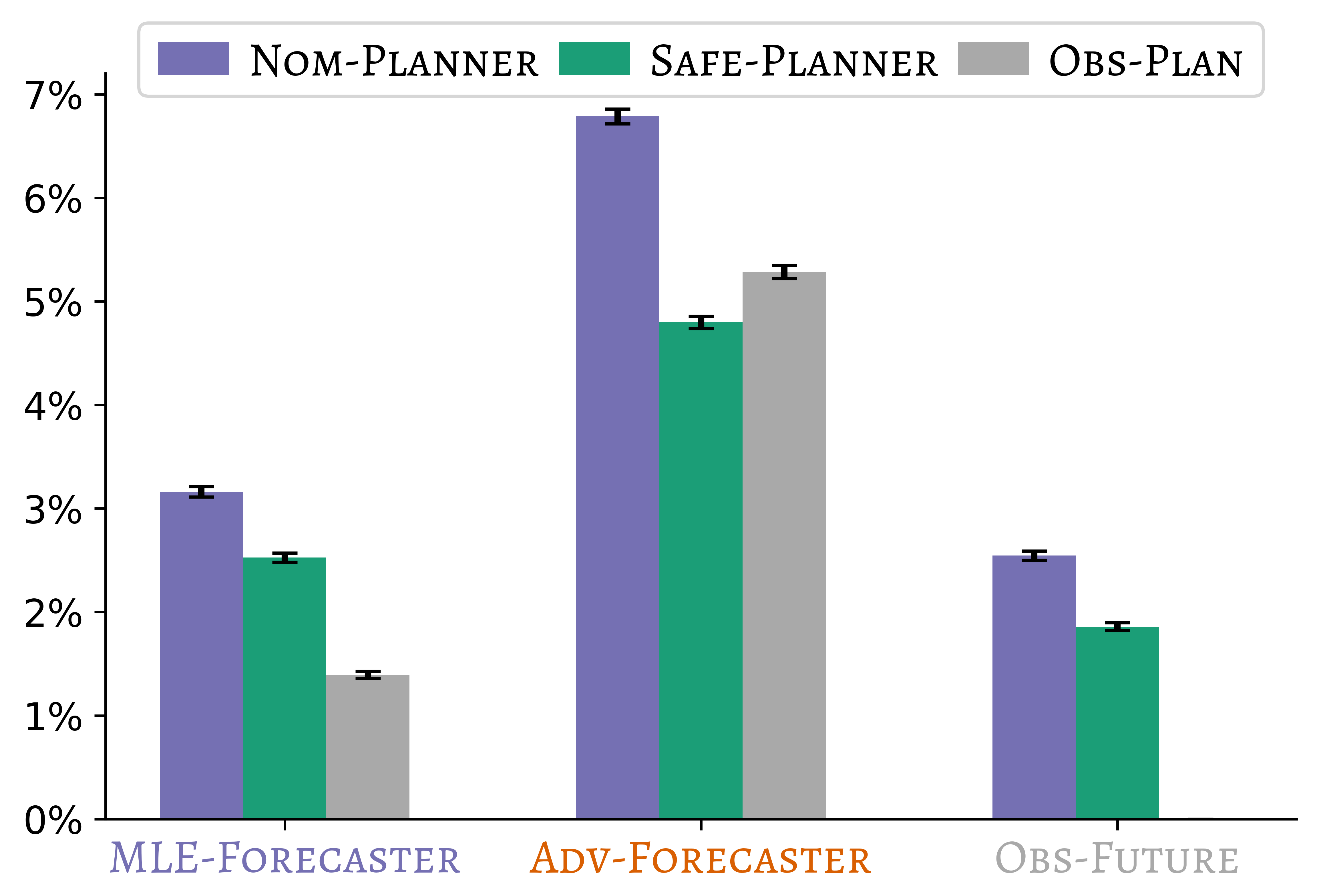}
    \caption{\small \textbf{CrowdNav} - CHOMP Cost (left) and Collision Rates (right)}
    \label{fig:task_overview_a}
    \end{subfigure}
    \begin{subfigure}[b]{.49\textwidth}
    \centering
    \includegraphics[width=.49\textwidth]{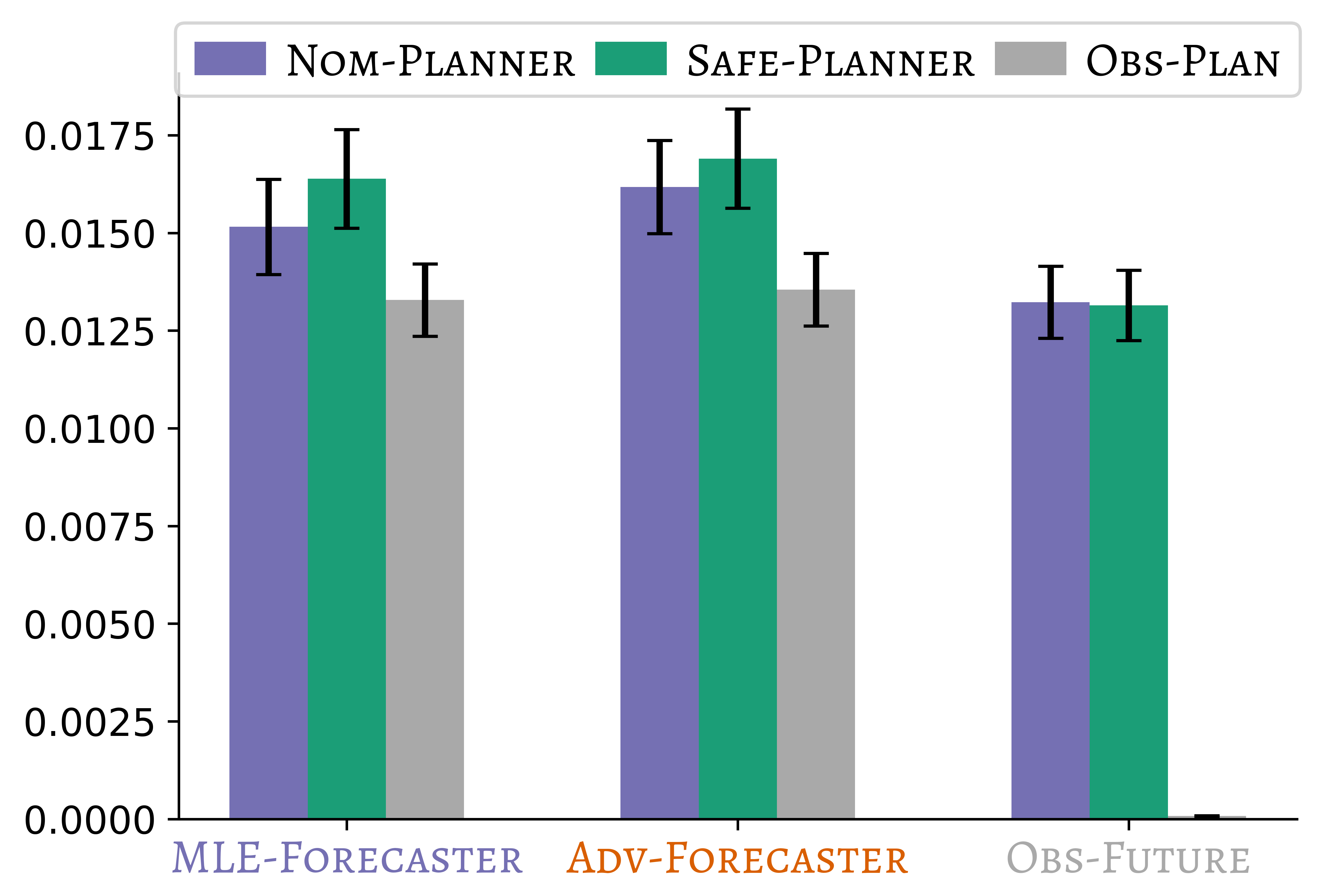}
    \includegraphics[width=.49\textwidth]{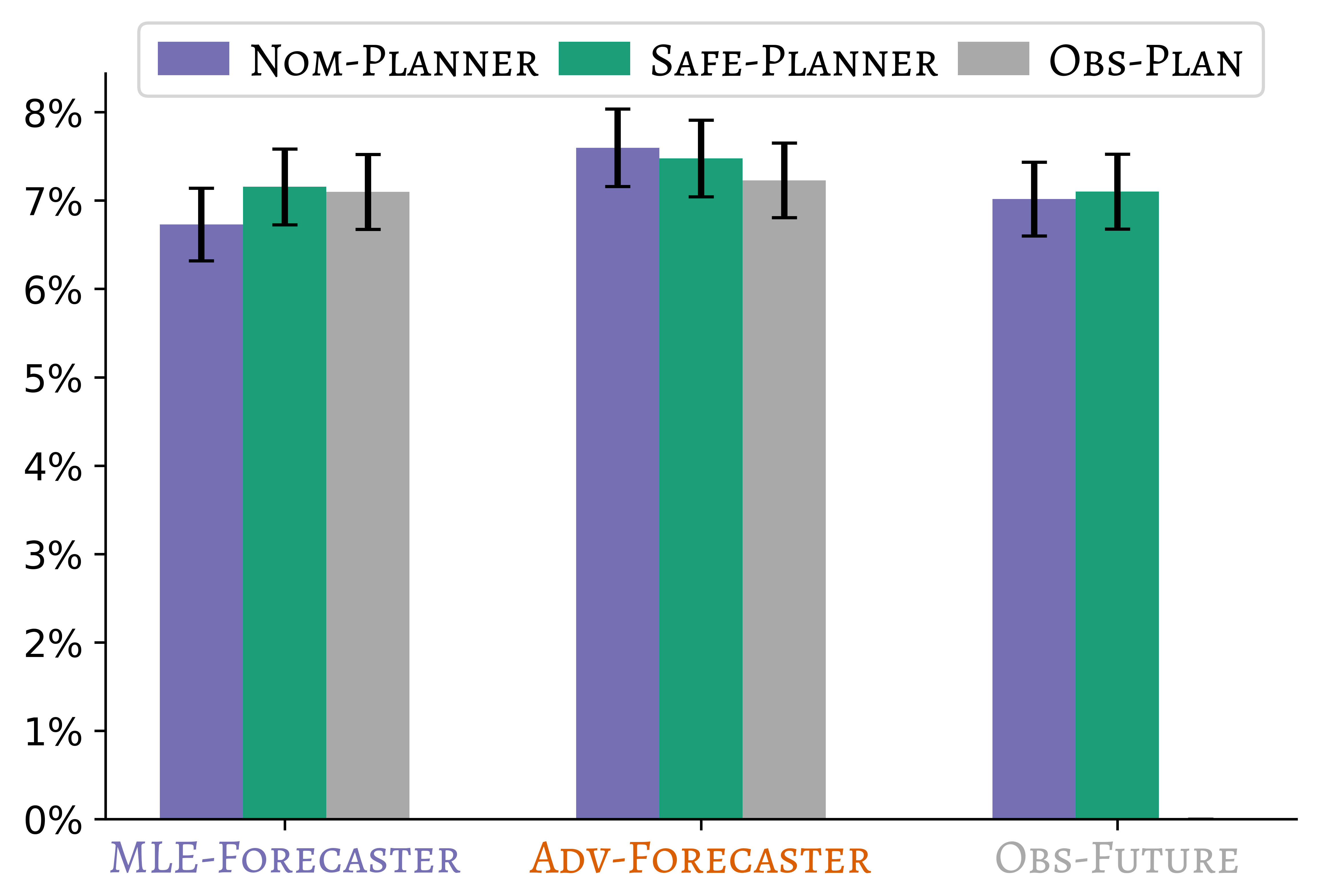}
    \caption{\small \textbf{ETH-UCY} - CHOMP Cost (left) and Collision Rates (right)}
    \label{fig:task_overview_b}
    \end{subfigure}
  \caption{Evaluation of CHOMP costs (Eq \ref{eq:chomp_cost}) and collision rates for different planner and forecaster combinations.}
  \vspace{-2mm}
  \label{fig:costs}
\end{figure*}

%% file: inputs/experiments/crowdnav.tex
\subsubsection{\textbf{\textsc{CrowdNav}} \cite{chen2019crowd}} This is an open-source simulator where a robot has to move forward in the presence of other humans. The humans in the scene move toward their respective goal locations and are simulated using the ORCA \cite{Berg2011ReciprocalNC} algorithm. We are interested in the non-compliant setting of the simulator, in which the five humans in the scene are unresponsive, i.e., they ignore robot motion. To navigate the scene, the robot should be able to plan around the future movements of the humans. To generate expert trajectories, we utilize the reinforcement learning (RL) agent provided by \cite{chen2019crowd}, which uses a self-attention (SA) module to encode human-human and human-robot interactions. The SARL agent is trained using a reward function that manually encodes collision avoidance and goal-reaching behavior.





We collect a dataset of 5000 episodes of human-robot navigation using this SARL policy. Our models are trained on 50\% of the dataset and evaluated on the rest. While the SARL model architecture considers just the current state of the robot and humans to predict the robot's immediate action, we also use an LSTM-module to encode a history of 8 timesteps (2 seconds) for each agent. The predictions produced by the forecaster are given as input to the planning module along with the context and goal location of the robot. We output actions for each agent over a horizon of 8 timesteps.

%% file: inputs/experiments/eth_ucy.tex

\subsubsection{\textbf{The \textsc{ETH-UCY} Benchmark}} There are five different datasets of real-world pedestrian movements in the \textsc{ETH} \cite{Pellegrini2009YoullNW} and \textsc{UCY} \cite{Lerner2007CrowdsBE} benchmark. The scenarios in the dataset showcase a wide range of human-human interactions and are a standard benchmark in the field. The data is captured at a 2.5Hz frequency (0.4s timestep). For the forecasting task, 8 timesteps of the history (3.2s) are considered, and 12 timesteps of the future are to be predicted for each agent. For evaluation, our model is trained on 4 out of the 5 datasets and evaluated on the held-out dataset. 

We implement the Trajectron++ \cite{Salzmann2020TrajectronDT} model for the base configuration of our planner and forecaster. It is a state-of-the-art multimodal conditional variational autoencoder (CVAE) generative model that can produce dynamically feasible trajectories. While the original model produces a distribution of trajectories, we use deterministic forecasts for each agent for simplicity. To do this, we restrict the outputs of the model to a unimodal distribution for each agent's future motion. To calculate the collision avoidance cost function, we use the mean of this distribution.


%% file: inputs/experiments/results.tex
\subsection{Results and Analysis}

\begin{figure}[t!]
  \centering
  \begin{subfigure}[b]{0.99\columnwidth}
    \centering {\includegraphics[width=0.99 \linewidth]{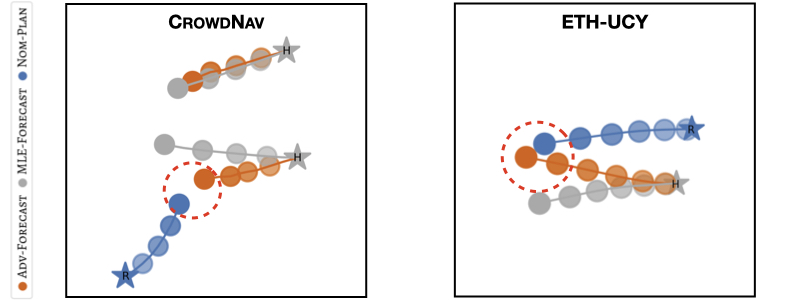}}
  \end{subfigure}
  \vspace{-4mm}
  \caption{\small{The \textsc{MLE-Forecaster} predicts the most likely futures for each human. \textsc{Nom-Planner} avoids collisions with the \textsc{MLE-Forecaster} but not with the \textsc{Adv-Forecaster}. Collisions are marked in red.}}
  \vspace{-2mm}
  \label{fig:o1}
\end{figure}

\textbf{\textit{O1.} \textsc{Adv-Forecaster} predicts more severe hazards than \textsc{MLE-Forecaster}.} In Fig. \ref{fig:o1}, we show examples of forecasts produced by the \textsc{Adv-Forecaster} that leads to collisions with the plans generated by the \textsc{Nom-Planner}. This is expected as the \textsc{Adv-Forecaster} is trained to increase the cost difference between generated plans and the observed trajectories in the dataset. On the other hand, \textsc{MLE-Forecaster} maximizes likelihood of observed motions and is unable to generate potential hazards that render the generated plans unsafe. Fig \ref{fig:costs} shows that the cost (Eq \ref{eq:chomp_cost}) of plans is significantly higher when evaluated with the adversarial forecasts compared with the MLE-Forecasts or the observed futures in the \textsc{CrowdNav} environment and the ETH-UCY benchmark.


\begin{figure}[t!]
  \begin{subfigure}[b]{0.99\columnwidth}
    \centering {\includegraphics[width=0.99 \linewidth]{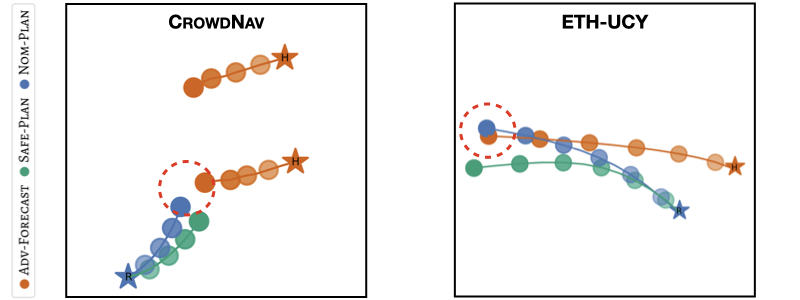}}
    \label{fig:2}
  \end{subfigure}
  \vspace{-4mm}
  \caption{\small{The \textsc{Safe-Planner} plans around the \textsc{Adv-Forecasts} leading to safe motions, whereas the \textsc{Nom-Planner} collides with the adversarial forecasts. Collisions are marked in red.}}
  \label{fig:o2}
  \vspace{-7mm}
\end{figure}
\textbf{{\textit{O2.}} \textsc{Safe-Planner} guards against rare events better than \textsc{Nom-Planner}. }
Fig \ref{fig:o2} shows scenarios where the \textsc{Nom-Planner} is in collision with the forecasts produced by the \textsc{Adv-Forecaster} as it does not consider the possibility of adverse events. Since the \textsc{Safe-Planner} is trained to minimize the cost difference with the adversarial forecasts, its plans are safe with respect to the \textsc{Adv-Forecaster}. It naturally encodes behavior that guards against rare events in the dataset. Fig \ref{fig:costs} shows that the \textsc{Safe-Planner} has lower costs than the \textsc{Nom-Planner} when evaluated against the observed futures of humans. We also observe lower costs and collision rates for the \textsc{Safe-Planner} compared to the \textsc{Nom-Planner} when tested against \textsc{MLE-Forecasts} and the \textsc{Adv-Forecaster} on the \textsc{CrowdNav} simulator. 



\begin{figure}
  \centering
  \begin{subfigure}[b]{0.99\columnwidth}
    \centering {\includegraphics[width=0.99 \linewidth]{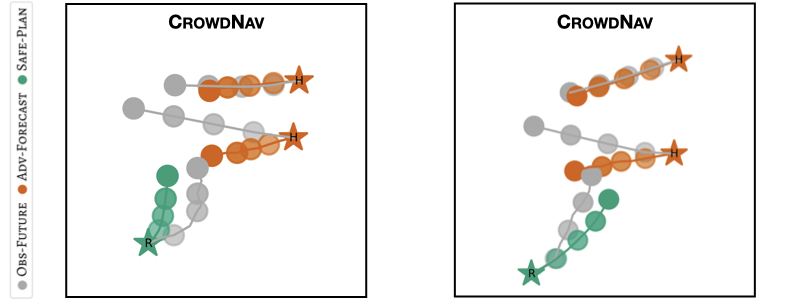}}
    \label{fig:2}
  \end{subfigure}
  \vspace{-4mm}
    \caption{\small{(\textbf{\textsc{CrowdNav}}) When \textsc{Adv-Forecaster} and \textsc{Safe-Planner} deviate from the ground truth predictions, they produce \textit{alternate plausible trajectories}. The forecasts produced by the \textsc{Adv-Forecast} represent risky futures. The \textsc{Safe-Planner} conservatively guards against possible rare events. }}
  \label{fig:o3}
  \vspace{-8mm}
\end{figure}

\textbf{\textit{O3.} \textsc{Safe-Planner} and \textsc{Adv-Forecaster} produce plausible trajectories even with higher tracking errors. }
Both the \textsc{Safe-Planner} and \textsc{Adv-Forecaster} are trained with the primary objective of optimizing cost difference. But to do so, they have to deviate from ground-truth observations. Table \ref{tab:errors} shows that their  average displacement error (ADE) and final displacement error (FDE) 
 is slightly higher. However, we observe that the trajectories generated by the models are generally plausible. Fig. \ref{fig:o3} shows scenarios in the \textsc{CrowdNav} simulator where they both deviate from ground truth trajectories but are still quite plausible counterfactuals that the robot should guard against. 



\input{inputs/tables/errors}
\vspace{-4mm}

%% file: inputs/tables/errors.tex
\begin{table}[t!]
\caption{\small{We compare the Average Displacement Error (ADE) and Final Displacement Error (FDE) of the predictions motions by our different planners/forecasters on the testing splits of the \textsc{ETH-UCY} benchmark and the \textsc{CrowdNav} simulator.}}
\label{tab:errors}
\resizebox{\linewidth}{!}{
\begin{tblr}{
  column{1} = {c},
  column{2} = {c},
  column{3} = {c},
  column{4} = {c=0}{0.00\linewidth},
  column{5} = {c},
  column{6} = {c},
  column{7} = {c}{0.00\linewidth},
  column{8} = {c},
  column{9} = {c},
  column{10} = {c}{0.00\linewidth},
  column{11} = {c},
  column{12} = {c},
  cell{1}{2} = {c=2}{},
  cell{1}{3} = {c=2}{0.03\linewidth},
  cell{1}{5} = {c=2}{},
  cell{1}{8} = {c=2}{},
  cell{1}{11} = {c=2}{},
  hline{1,5} = {-}{0.15em},
  hline{3} = {-}{0.05em},
  hline{2} = {2-3, 5-6, 8-9, 11-12}{0.05em},
}
            & \textsc{MLE-Forecaster} &       &  & \textsc{Adv-Forecaster} &       &  & \textsc{Nom-Planner} &       &  & \textsc{Safe-Planner} &       \\
            & ADE            & FDE   &  & ADE            & FDE   &  & ADE         & FDE   &  & ADE          & FDE   \\
\textsc{ETH-UCY} & 0.387          & 0.947 &  & 0.405          & 0.950 &  & 0.387       & 0.947 &  & 0.391        & 0.956 \\
\textsc{CrowdNav}    & 0.268          & 0.371 &  & 0.274          & 0.383 &  & 0.184       & 0.268 &  & 0.193        & 0.283 
\end{tblr}
}
\end{table}

%% file: inputs/discussion.tex
\vspace{3mm}
\section{Discussion and Limitations}
\vspace{-1mm}
This paper introduces a novel game-theoretic framework that addresses joint forecasting and planning. We discuss the \textbf{pitfalls of MLE forecasting} that only focus on maximizing the likelihood of observed human motion. Instead, we produce \textbf{adversarial counterfactuals} by optimizing the performance difference between generated plans and observed demonstrations, considering the predictions made by our learned forecaster. In response, our framework can \textbf{guard against rare but risky events} by generating plans that are safe with respect to the adversary.

There are some limitations to our approach. We observed larger error ranges on the ETH-UCY dataset. This is likely because real-world pedestrian datasets contain significant noise in estimation and a wide variety of behaviors, making it difficult to model human behavior accurately. In future work, we will extend our framework to consider multi-modal distributions of plans and forecasts. On-policy evaluation of our framework in scenarios where humans suddenly change their goals is another promising direction.
